\documentclass[letterpaper, 10 pt, conference]{ieeeconf}  

\IEEEoverridecommandlockouts                              

\usepackage{graphics} 
\usepackage{epsfig} 
\usepackage{multirow}
\usepackage{colortbl}
\usepackage[top=19truemm,bottom=19truemm,left=19truemm,right=19truemm]{geometry}
\usepackage{amssymb}
\usepackage{amsmath}
\usepackage{color}
\usepackage{booktabs}
\usepackage{url}

\title{\LARGE \bf
Imitation Learning Based on Bilateral Control\\for Human--Robot Cooperation
}

\author{Ayumu Sasagawa$^{1}$, Kazuki Fujimoto$^{2}$, Sho Sakaino$^{3}$, and Toshiaki Tsuji$^{4}$
\thanks{$^{1}$Ayumu Sasagawa is a student with the Graduate School of Science and Engineering,
Saitama University, 255 Shimo-Okubo, Sakura-ku, Saitama City, Saitama 338-8570, Japan
       {\tt\small email: a.sasagawa.997@ms.saitama-u.ac.jp}}%
\thanks{$^{2}$Kazuki Fujimoto is a student with the Graduate School of Science and Engineering,
Saitama University, 255 Shimo-Okubo, Sakura-ku, Saitama City, Saitama 338-8570, Japan
        {\tt\small email: k.fujimoto.423@ms.saitama-u.ac.jp}}%
\thanks{$^{4}$Sho Sakaino is with the Graduate School of Systems and Information Engineering, University of Tsukuba, 1-1-1 Tennodai, Tsukuba, Ibaraki 305-8577, Japan and the JST PRESTO
        {\tt\small email: sakaino@iit.tsukuba.ac.jp}}%
\thanks{$^{3}$Toshiaki Tsuji is with the Graduate School of Science and Engineering,
Saitama University, 255 Shimo-Okubo, Sakura-ku, Saitama City, Saitama 338-8570, Japan
        {\tt\small email: tsuji@ees.saitama-u.ac.jp}}%
        }
\begin{document}
\maketitle
\thispagestyle{empty}
\pagestyle{empty}
\begin{abstract}
  Robots are required to autonomously respond to changing situations.
  Imitation learning is a promising candidate for achieving generalization performance, and extensive results have been demonstrated in object manipulation.
  However, cooperative work between humans and robots is still a challenging issue because robots must control dynamic interactions among themselves, humans, and objects.
  Furthermore, it is difficult to follow subtle perturbations that may occur among coworkers.
  In this study, we find that cooperative work can be accomplished by imitation learning using bilateral control.
  Thanks to bilateral control, which can extract response values and command values independently, human skills to control dynamic interactions can be extracted.
  Then, the task of serving food is considered.
  The experimental results clearly demonstrate the importance of force control, and the dynamic interactions can be controlled by the inferred action force.
\end{abstract}
\section{INTRODUCTION}
Robots are expected to operate alongside humans in real environments like homes, factories, {\it etc}. 
In such environments, situations in which robots collaborate with humans are supposed, in addition to the case where the robot operates alone.
To perform tasks in such environments, robots are required to flexibly respond to environmental changes and adapt to unknown objects and situations.
Furthermore, when cooperating with humans, robots must also consider interactions with humans.
To deal with continually changing conditions and dynamic interactions with target objects and humans is still a challenging issue for robots.\par
To address these problems, considerable research on robot motion planning by machine learning has been reported \cite{end-to-end} \cite{machine learning1}. 
Levine {\it et al}. succeeded in grasping various objects using reinforcement learning based on end-to-end learning \cite{google}.
Also recently, methods called ``imitation learning'' or ``learning from demonstration'' have been reported to imitate human skills from demonstrations \cite{ogata} \cite{osa}. Most of these studies generated robotic trajectories based on position information \cite{one-shot}. 
However, they failed to demonstrate high performance.
Note that, every robot task is described as a combination of position and force controllers with adequate transformation of variables \cite{analysis}.\par
Therefore, in imitation learning for object manipulation, it is desirable to consider force information in addition to position information.
Some studies on imitation learning using force information have been reported \cite{iit1}--\cite{force}. 
However, tasks with fast motion have not yet been verified in these studies.
The reason for this is detailed later in this section.
Alternatively, several methods have been proposed for such cooperative work.
Calinon {\it et al}. validated the performance of the transportation task as cooperative work through demonstrations based on 2ch bilateral control \cite{cooperation2ch}. In this research, force control was not implemented in the robot, and it is expected that it is difficult to handle objects with undefined shape.
Rozo {\it et al}. also tried the transporting task exploiting force information to adapt to changes in human motion \cite{cooperationIIT}. 
In this study, the object was limited to a box, therefore the performance of objects with undefined shape and flexible objects was not verified. Also, the verification of fast motions and environmental changes was not sufficient.
In addition, Batinica {\it et al}. reported a cooperative operation method for a bimanual robot that can cope with external perturbations without an explicit dynamic model of the task \cite{cooperationBimanual}. However, cooperative work with humans has not been verified.\par
\begin{figure}[t]
  \centering
  \includegraphics[width=7cm]{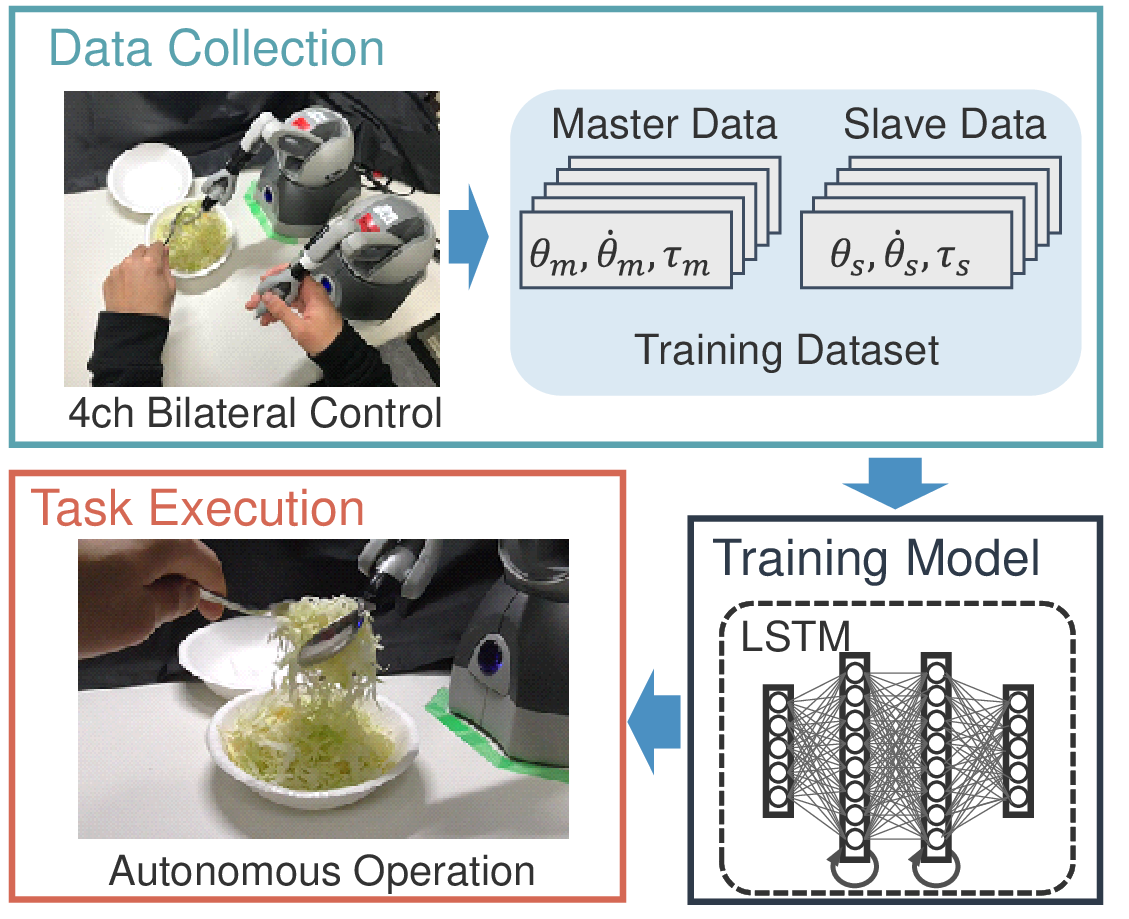}
  \caption{System overview of the proposed method}
  \label{fig:system}
\end{figure}
Hence, in this paper, the performance of scooping and transportation tasks via human-robot cooperation was verified.
The performance was evaluated by the success rate of the task.
The task was validated for various objects including flexible objects and objects with undefined shape. 
The proposed method was a method based on 4ch bilateral control.
Bilateral control is a remote control technology using a master robot and a slave robot \cite{bilate} \cite{symmetric_bilate}.
Previously, Adachi {\it et al}. have established a method of imitation learning for object manipulation that implements position and force control using 4ch bilateral control \cite{adachi}.
Owing to 4ch bilateral control, this method has high generalization ability.
By imitating human skills that include force information, they succeeded in tasks requiring force adjustment.
Another notable feature of this method is that the robots can move as fast as humans.\par
A system overview of the proposed method is shown in Fig.~\ref{fig:system}. 
The proposed method focused on the realization of tasks requiring force information and fast motion considering dynamic interactions among the robot, environment, and human.
Here, the important points of the method based on 4ch bilateral control for cooperation are detailed as follows.
\subsubsection{Command values of the slave must be predicted}\label{subsubsec:command}
\begin{figure}[t]
  \centering
  \includegraphics[width=8cm]{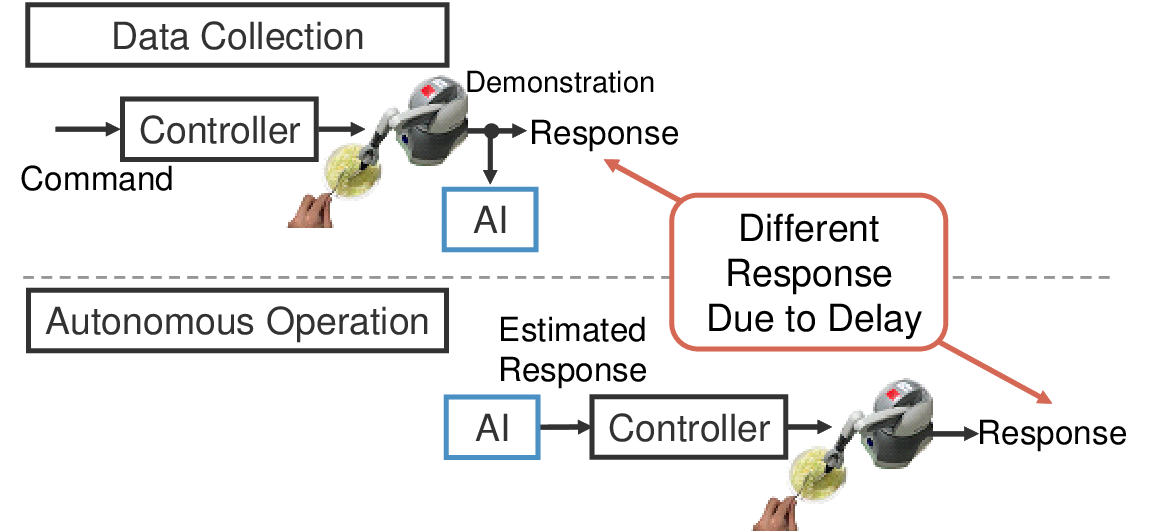}
  \caption{Overview of general imitation learning. The response values measured during the demonstration are used as the command values of the controller during autonomous operation. Even if the response values can be predicted perfectly as the command values, the actual response values of the robot are delayed from the command values. Therefore, when the robot actually operates, the response values is different from the estimated response values.}
  \label{fig:DelayConv}
\end{figure}
\begin{figure}[t]
  \centering
  \includegraphics[width=8cm]{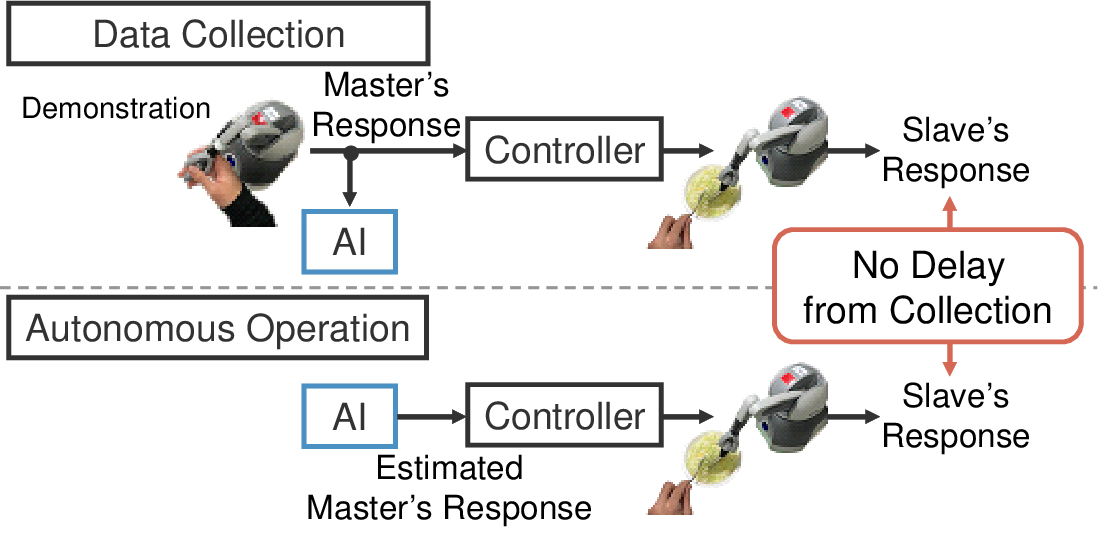}
  \caption{Overview of the proposed method. The master response values are used as the slave command values. As the figure illustrates, the command values to the response values are configured by the same system during data collection and autonomous operation. Therefore, the delay from the command values to the response values is exactly the same during data collection and autonomous operation.}
  \label{fig:DelayProp}
\end{figure}
At first, the general method for imitation learning \cite{iit1}--\cite{force}, as show in Fig.~\ref{fig:DelayConv}, is explained.
They collected response values during the demonstration and gave it as command values of controllers because command values cannot be directly measured. 
Then, controllers are implicitly assumed to have sufficiently high performance such that transfer functions between command and response values are equal to 1. 
As this is obviously difficult, this method can operate only in slow motion where dynamics are negligibly small and transfer functions can be 1.\par
In order to achieve fast motion, we need to collect command values considering dynamic interactions among the robot, environment, and operator. 
In other words, we need to independently measure command and response values because the human skills required to handle the dynamic interactions and to achieve fast manipulation are included in
the command values.
The proposed method using bilateral control is shown in Fig.~\ref{fig:DelayProp}.
When a slave robot is teleoperated by a master robot, the response of the master is given as a command to the slave. 
As a result, both command and response values can be measured.
Therefore, the command values of the slave can be predicted during autonomous operation.
As shown in Fig.~\ref{fig:DelayProp}, the relationship from the command values to the response values is the same during data collection and autonomous operation.
It is not necessary to assume transfer functions to be 1 because the control delays and delays due to physical interaction that occurs during data collection similarly occurs during autonomous operation.
From the above, the proposed method can achieve fast motion.
The robot can move as fast as the robot was controlled by 4ch bilateral control. The maximum speed is limited by the control performance of 4ch bilateral control.
\subsubsection{Slave must consit of position and force control}\label{subsubsec:force control}
As mentioned above, the robot motion consists of position and force controls \cite{analysis}.
Although the necessity of force information in imitation learning was mentioned, in order to realize it, both position and force controls must be implemented in the slave.
\subsubsection{The control system must be the same during data collection and autonomous operation}\label{subsubsec:same system}
During the demonstration, humans collect data considering both the control delays and delays due to physical interactions occur, {\it i.e.} humans compensate for these delays.
If the control system changes during data collection and autonomous operation, such human compensation skills will be lost.
Therefore, the control system must be the same during each phase.\par

Considering the above three points, the proposed method was designed to reproduce the bilateral control.
In the proposed method, 4ch bilateral control \cite{bilate} was adopted from among various types of bilateral control because 4ch bilateral control has the highest performance and excellent operability, and position/force controllers are implemented in the slave.

In this paper, experiments were validated on a variety of objects, including flexible objects and objects with undefined shape.
Also, it was verified assuming that there was an error in the height at which the target object was placed and in the kinematics of the robot, and the case in which the human motion was different from the demonstration.
The effectiveness of the proposed method in human--robot cooperation was verified through comparative experiments using several methods.
In the comparative experiments, the performance of each method was evaluated by the success rate of the task.
The contribution of force control and the effectiveness of employing bilateral control to control the dynamic interactions were clarified through these comparative experiments.
Note that the proposed method does not use visual information.
In order to perform the tasks we verified, the adjustment of force was important, and the robot needs to consider dynamic interactions with the human and environments.
It is difficult to perform the tasks based only on visual and position information because the dynamic interactions cannot be measured by the visual and position information.
Therefore, this study addressed the problem using force control and our proposed framework.\par

This paper has contributions as a method of imitation learning for human-robot cooperation, because it can be realized in the same framework as a conventional method \cite{adachi} without explicitly considering dynamic interactions. 
In addition, the framework based on 4ch bilateral control makes a great contribution in the field of imitation learning because it can realize tasks requiring adjustment of force and fast motion.\par
This paper is organized as follows. Section~\ref{sec:CONTROL SYSTEM} explains the control system including bilateral control. Section~\ref{sec:system} introduces the proposed method of imitation learning.
This section describes the data collection method using bilateral control, learning method, and task execution. Section~\ref{sec:experiment} describes the experiments and presents the results. 
Section~\ref{sec:conclusion} concludes this study and discusses future works.
\section{CONTROL SYSTEM}
\label{sec:CONTROL SYSTEM}
\subsection{Manipulator}\label{subsec:manipulater}
\begin{figure}[t]
  \centering
  \includegraphics[width=7cm]{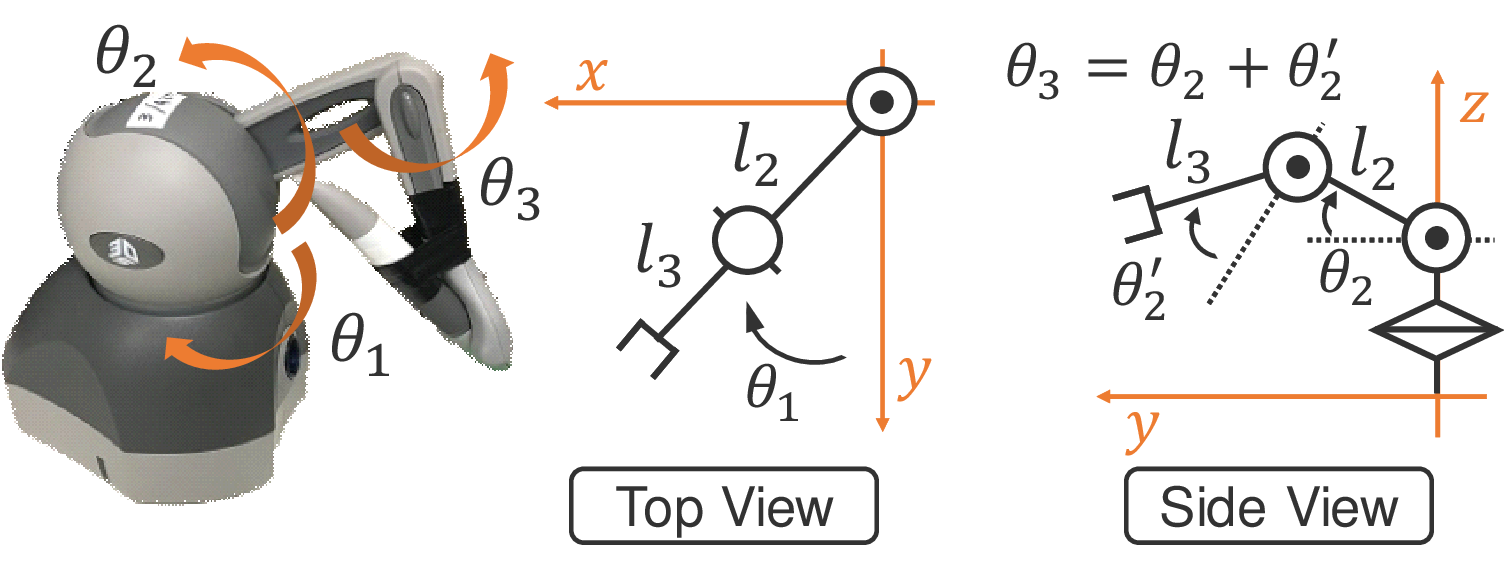}
  \caption{Touch${}^\text{\tiny TM}$ USB haptic device}
  \label{fig:PHANToM}
\end{figure}
\begin{figure}[t]
  \centering
  \includegraphics[width=6cm]{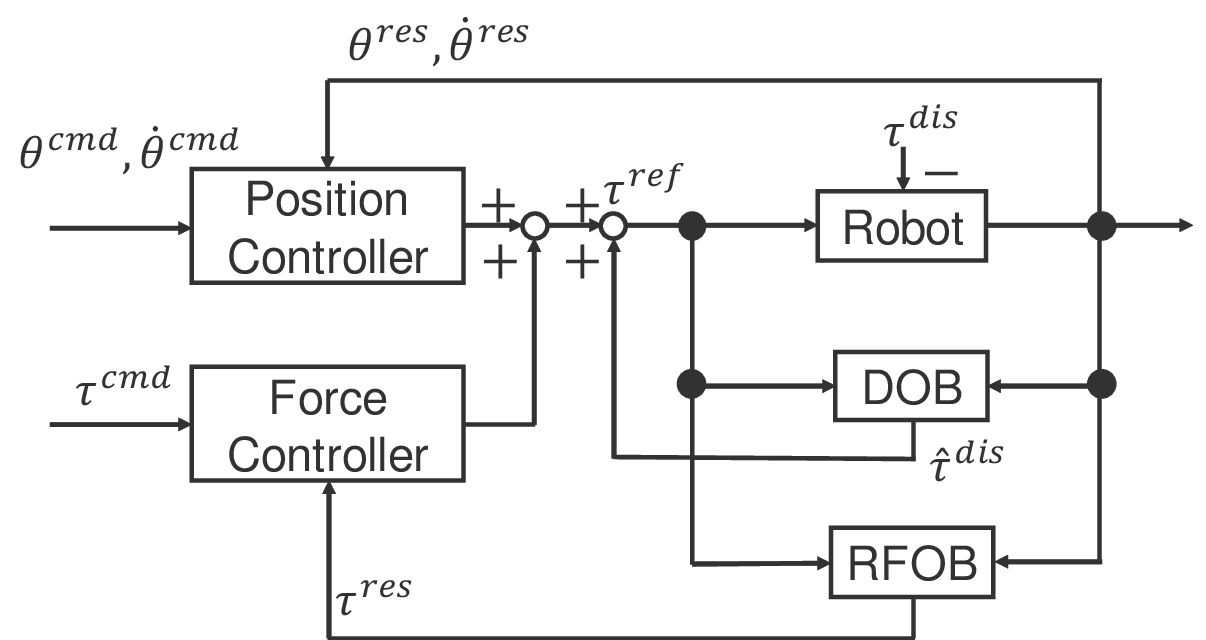}
  \caption{Controller}
  \label{fig:controller}
\end{figure}
We used two Touch${}^\text{\tiny TM}$ USB haptic devices manufactured by 3D Systems (Fig.~\ref{fig:PHANToM}).
Joint angles of the manipulator are shown in the center and right side of Fig.~\ref{fig:PHANToM}.
A diagram of the manipulator system is shown in Fig.~\ref{fig:controller}.
Here, $\theta$, $\dot{\theta}$, and $\tau$ represent joint angles, angular velocity, and robot torque, respectively. The superscripts res, ref, and cmd indicate response, reference, and command values, respectively.
The controller was composed of the combination of position and force controllers; the position controller consisted of proportional and derivative controllers while the force controller included a proportional controller.
These manipulators measured the angle $\theta^{res}$ at each joint. Angular velocity $\dot{\theta}^{res}$ was calculated using a pseudo-derivative and disturbance torque $\tau^{dis}$ was estimated by a disturbance observer (DOB) \cite{dob}. Moreover, the reaction force $\tau^{res}$ was calculated by an RFOB \cite{rfob}.
In this study, robots were operated with a 1~ms control cycle.
\subsection{4ch Bilateral Control}\label{subsec:bilate}
\begin{table}[t]
  \tabcolsep = 3pt
  \begin{center}
  \caption{Gains and identified system parameters \protect\linebreak for robot controller}
  \label{table:parameter}
  \begin{tabular}{lccc}
      \hline\hline
      & Parameter & Master & Slave \\
      \hline
      \arrayrulecolor{black}
      $J_{1}$            & inertia ($\theta_{1}$) [$\rm{mkgm^2}$]   & 2.92 & 2.71 \\
      $J_{2}$            & inertia ($\theta_{2}$) [$\rm{mkgm^2}$]   & 3.44 & 3.08 \\
      $J_{3}$            & inertia ($\theta_{3}$) [$\rm{mkgm^2}$]   & 1.05 & 1.07 \\
      $g_{c1}$                       & Gravity compensation coefficient 1 [$\rm{mNm}$]   & 124 & 106 \\
      $g_{c2}$                       & Gravity compensation coefficient 2 [$\rm{mNm}$]   & 81.6 & 96.3 \\
      $g_{c3}$                       & Gravity compensation coefficient 3 [$\rm{mNm}$]   & 81.6 & 85.1 \\
      $D$                         & Friction compensation coefficient[$\rm{mkgm^2/s}$]   & 6.87 & 12.0 \\
      \hline
      $K_p$                       & Position feedback gain              & \multicolumn{2}{c}{121.0} \\        
      $K_d$                       & Velocity feedback gain                & \multicolumn{2}{c}{22.0} \\
      $K_f$                       & Force feedback gain              & \multicolumn{2}{c}{1.0} \\
      \hline\hline
  \end{tabular}
  \end{center}
\end{table}

\begin{figure}[t]
  \centering
  \includegraphics[width=6cm]{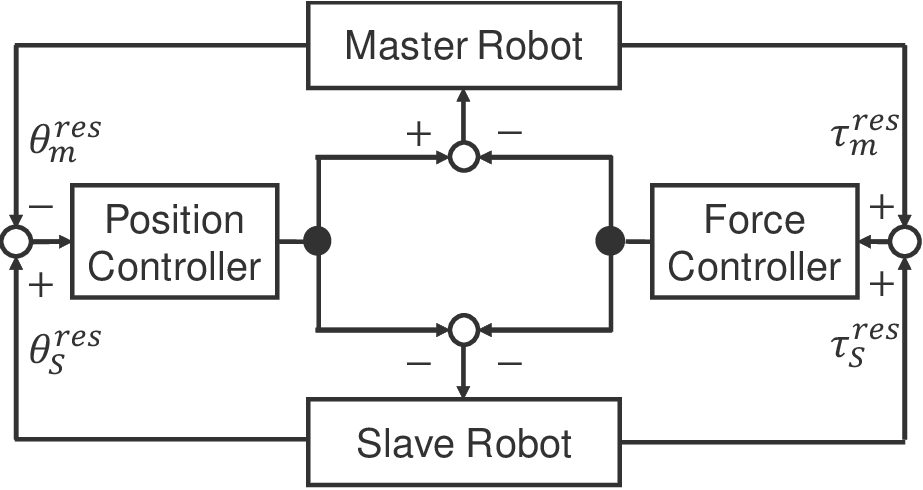}
  \caption{4ch bilateral controller}
  \label{fig:bilate}
\end{figure}

In the proposed method, 4ch bilateral control was used for collecting training datasets. 
4ch bilateral control is a remote-control system using two robots, a master and a slave.
Because the positions of the two robots are synchronized and the forces of each other are fed back, the operator that operates master can feel the interactions of the slave.
The control target of 4ch bilateral control is shown in (\ref{eq1}) and (\ref{eq2}), and a block diagram that satisfies (\ref{eq1}) and (\ref{eq2}) is shown in Fig.~\ref{fig:bilate} \cite{bilate}.
\begin{eqnarray}
  \theta^{res}_{m}-\theta^{res}_{s} & = & 0 \label{eq1} \\
  \tau^{res}_{m}+\tau^{res}_{s} & = & 0. \label{eq2}
\end{eqnarray}
Subscripts $m$ and $s$ indicate the master robot and slave robot, respectively.
The torque reference values $\tau^{ref}$ of each robot are defined as follows:
\begin{eqnarray}
  \tau^{ref}_{m} & = & - C_{p} (\theta^{res}_{m} - \theta^{res}_{s}) - C_{f} (\tau^{res}_{m}+\tau^{res}_{s})\label{eq3} \\
  \tau^{ref}_{s} & = &   C_{p} (\theta^{res}_{m} - \theta^{res}_{s}) - C_{f} (\tau^{res}_{m}+\tau^{res}_{s}).\label{eq4}
\end{eqnarray}
Here, $C_{p}$ and $C_{f}$ represent a position controller and a force controller, respectively.
$C_{p}$ is expressed as $C_{p} = \frac{J}{2}(K_{p} + K_{d}s)$ using the inertia $J$, position control gain $K_p$, and speed control gain $K_d$, and $C_f$ is expressed as $C_{f} = \frac{1}{2}K_{f}$ using force control gain $K_f$. Here, $s$ is the Laplace operator. The value of each gain used in the experiment is shown in Table~\ref{table:parameter}.
In this study, the reaction torques at each joint were calculated by RFOB as detailed in Section~\ref{subsec:rfob}.
The implementation cost was reduced as a force sensor was not used in this method.
\subsection{The system identification for control}\label{subsec:rfob}
Using the inertia $J$, the friction compensation coefficient $D$, and the gravity compensation coefficient $g_c$, the robot dynamics can be expressed by the following equation.
\begin{eqnarray}
  J_{1}\ddot{\theta}^{res}_{1} & = & \tau^{ref}_{1} - \tau^{dis}_{1} - D \dot{\theta^{res}_{1}}\label{eq:dob1} \\
  J_{2}\ddot{\theta}^{res}_{2} & = & \tau^{ref}_{2} - \tau^{dis}_{2} - g_{c1} \cos{\theta^{res}_{2}} - g_{c2} \sin{\theta^{res}_{3}}\label{eq:dob2} \\
  J_{3}\ddot{\theta}^{res}_{3} & = & \tau^{ref}_{3} - \tau^{dis}_{3} - g_{c3} \sin{\theta^{res}_{3}}.\label{eq:dob3}
\end{eqnarray}
Here, the numbers in subscript represent each joint of the robot. Additionally, the off-diagonal term of the inertia matrix was ignored because it was negligibly small. The physical parameters were identified based on \cite{yamazaki}. The parameters $D$ and $g_c$ were identified in free motion assuming $\tau^{dis} = 0$. The disturbance torque $\hat{\tau}^{dis}$ that is actually estimated by the DOB is shown as follow:

\begin{eqnarray}
  \hat{\tau}^{dis} & = & \tau^{ref} - J\ddot{\theta}^{res}.\label{eq:dob4}
\end{eqnarray}
The torque response values of each joint were calculated by RFOB expressed as follows:

\begin{eqnarray}
  \tau^{res}_{1} & = & \hat{\tau}^{dis}_{1} - D \dot{\theta^{res}_{1}}\label{eq:dob5} \\
  \tau^{res}_{2} & = & \hat{\tau}^{dis}_{2} - g_{c1} \cos{\theta^{res}_{2}} - g_{c2} \sin{\theta^{res}_{3}}\label{eq:dob6} \\
  \tau^{res}_{3} & = & \hat{\tau}^{dis}_{3} - g_{c3} \sin{\theta^{res}_{3}}.\label{eq:dob7}
\end{eqnarray}
Each identificated parameter used in the experiment is shown in Table~\ref{table:parameter}.

\section{SYSTEM FOR IMITATION LEARNING USING\\4CH BILATERAL CONTROL}
\label{sec:system}
This section explains the proposed approach.
The goal of this study is to enable a human and robot to perform tasks through cooperative work.
A robot autonomously performed cooperative work with a human by learning object manipulation.
The proposed imitation learning was executed in the following phases:
\begin{enumerate}
  \item Data collection phase
  \item Training neural network model phase
  \item Task execution phase. 
\end{enumerate}
The details of each phase are described below.
\subsection{Data Collection Phase}
The training datasets were collected using 4ch bilateral control.
4ch bilateral control was implemented in the master and slave robots.
A human operated the master robot directly with their hand.
Simultaneously, the slave robot performed tasks in the workspace.
Here, recorded motion data were angle, angular velocity, and torque response values of both the robots.
In the proposed method, the responses of the master robots are equivalent to the commands of the slave robot.
As the operator can recognize the control delays of the slave robot, the operator can modify the commands of the master robot to compensate for the control delays.
In other words, in the proposed method, the skills of humans to adapt to the control delays can also be extracted.
\subsection{Training Neural Network Model Phase}
\begin{figure}[t]
  \centering
  \includegraphics[width=8cm]{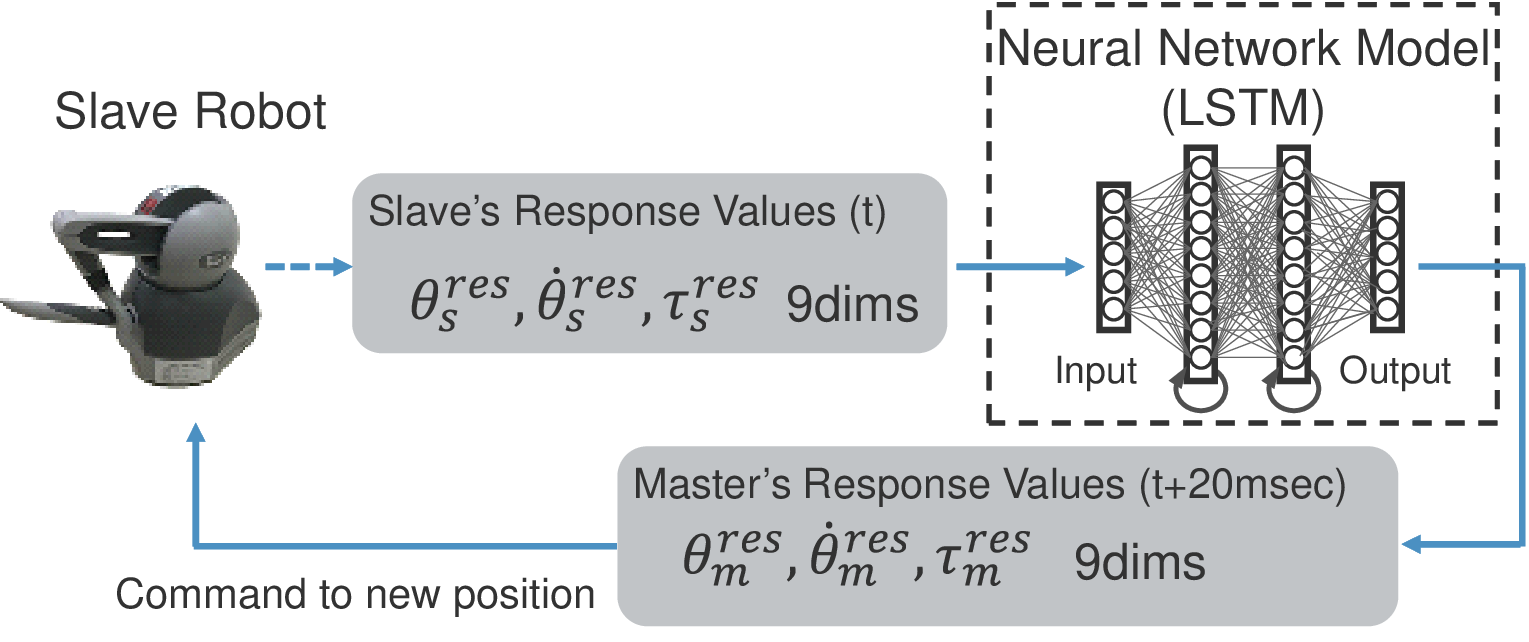}
  \caption{Construction of proposed neural network model. Input to long short-term memory (LSTM) is the angle, angular velocity, and torque response values of the slave robot.
  The input has nine-dimensions because the slave robot is a 3-DoF robot.
  Similarly, the output response values of the master robot have nine-dimensions. The LSTM model infers the states of the master robot 20~ms after the input is applied.}
  \label{fig:NNarc}
\end{figure}
The proposed neural network model is shown in Fig.~\ref{fig:NNarc}.
Training data collected by 4ch bilateral control were used to train the model.
In this study, robotic motion was realized by end-to-end learning. \par
A recurrent neural network (RNN) was adopted as the neural network that generated robotic motion.
RNN was suitable for the inference of the sequence data.
Recently, research on robotic motion generation using RNN has been reported \cite{rnn3}.
In this study, long short-term memory (LSTM), which is a type of RNN, was used. LSTM is a neural network that can handle long time series data as compared to RNN \cite{LSTM}.\par
\begin{figure}[t]
  \centering
  \includegraphics[width=8cm]{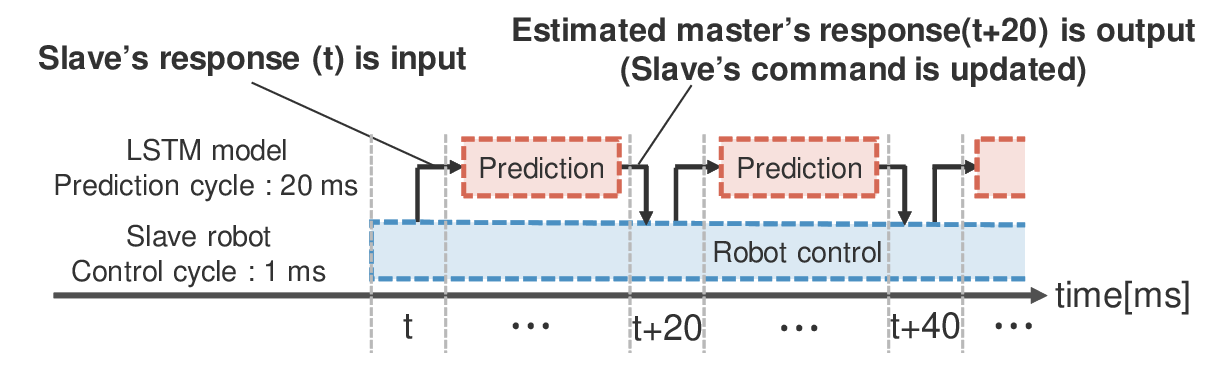}
  \caption{The relationship between the control cycle of the slave and the prediction cycle of the LSTM model. The model predicted the state of the master 20 msec later from the input, and predicted values were input as new command values.}
  \label{fig:cycleRelationship}
\end{figure}

The neural network model used in this study had two LSTM layers with unit sizes of 50, followed by a fully-connected layer.
The input--output responses of the LSTM are shown in Fig.~\ref{fig:NNarc}.
The input of the LSTM had nine-dimensions, including angle, angular velocity, and torque for each joint of the slave robot.
Similarly, the output also included the angle, angular velocity, and torque for each joint of the master robot.
The output inferred the states of the master robot 20~ms after the input was applied.
The relationship between the control cycle of the slave and the prediction cycle of the LSTM model during autonomous operation is shown visually in Fig.~\ref{fig:cycleRelationship}.
The model was designed to predict 20 ms later from input assuming that the robot motion would be generated online using the learned model during autonomous operation.
20ms was decided considering the time the model took to predict.
Because the model took 20 msec from input to output, the predicted values were output 20 ms later from input and that it was input to the robot as command values.
This implies that the command in the next step of the slave robot was inferred using LSTM.
Input--output data were normalized as preprocessing for learning in the same way as the conventional method \cite{adachi}. 
In the proposed model, the parameters of LSTM were updated according to the output errors.
The loss function was the mean square error of the inference values and training data values.
\subsection{Task Execution Phase}
In this phase, the robot performed tasks autonomously in cooperation with a human.
The robot measured sensor information in real-time and generated robotic motion using the trained model, {\it i.e.}, the slave robot worked using neural network inference.
While the robot was moving, angle, angular velocity, and torque of the slave robot were input to the LSTM model every 20~ms.
LSTM output the angle, angular velocity, and torque, and these data were the command values for the subsequent action to be performed by the robot.
Similar to the training prosess, the outputs of LSTM were obtained at 20~ms intervals.
\begin{figure}[t]
  \centering
  \includegraphics[width=8.2cm]{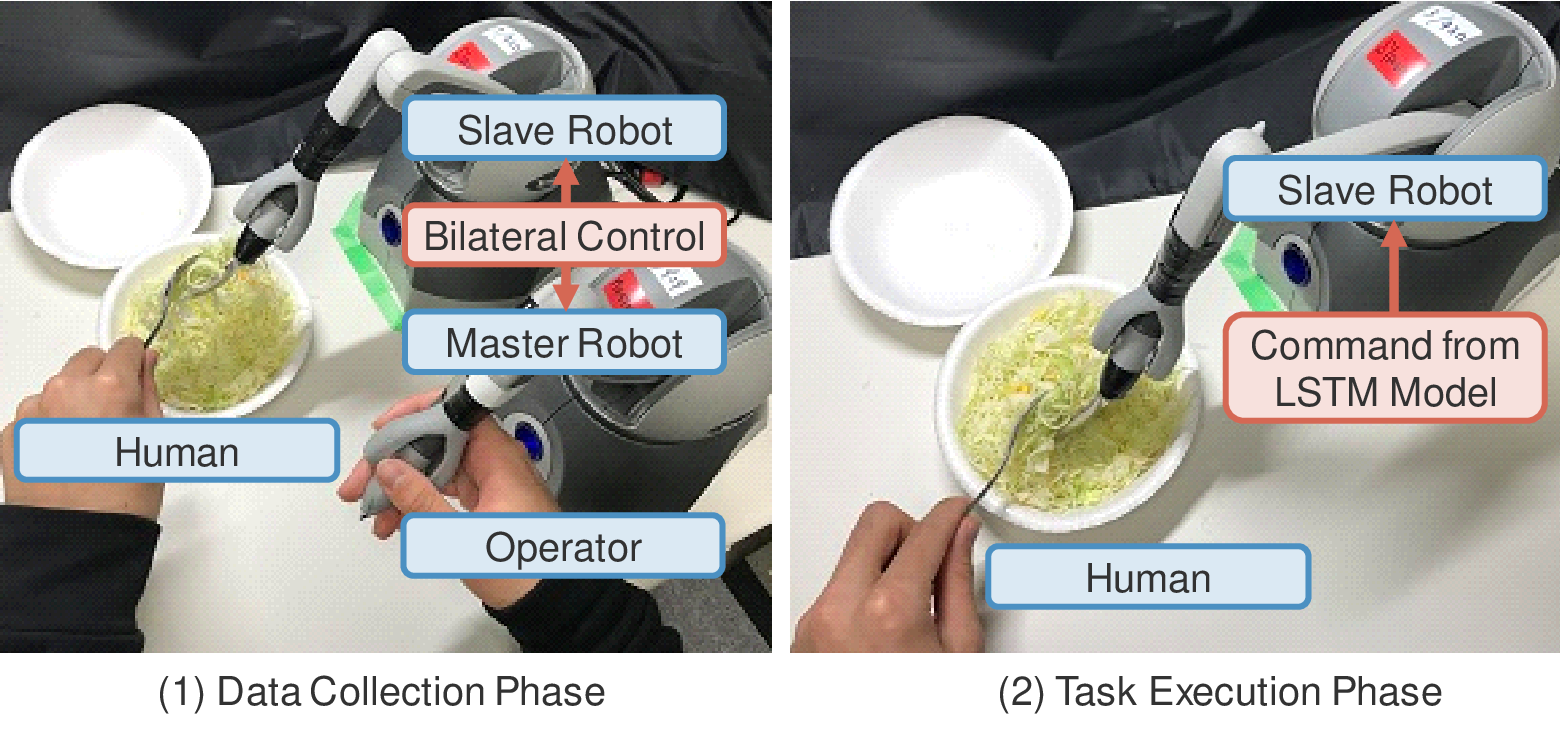}
  \caption{Experimental setup for data collection and task execution phases}
  \label{fig:experiment}
\end{figure}
\begin{figure}[t]
  \centering
  \begin{tabular}{c}
    \centering
    \includegraphics[width=7cm]{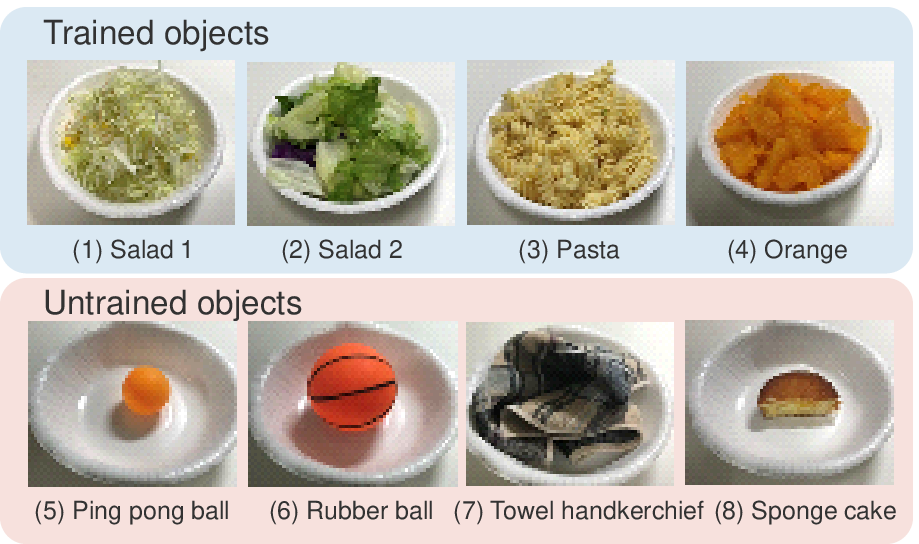}
  \end{tabular}
  \caption{Objects used in the experiment}
  \label{fig:food}
\end{figure}
\begin{figure}[t]
  \centering
  \includegraphics[width=6.5cm]{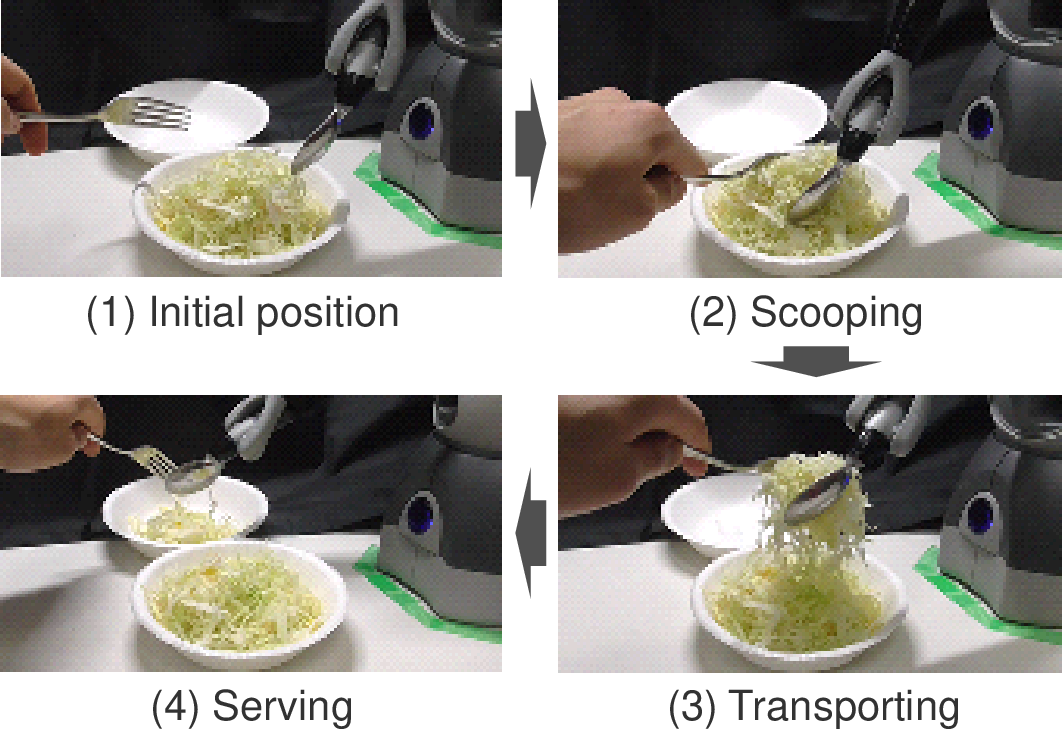}
  \caption{Task sequence of serving salad}
  \label{fig:salad_seq}
\end{figure}
\section{EXPERIMENT}
\label{sec:experiment}
\begin{table*}[t]
  \begin{center}
  \caption{Details of comparative experiment}
  \label{table:experiment}
  \begin{tabular}{|l||c|c|c|c|c|}
      \hline
      &                            & \multicolumn{2}{c|}{Controller} & \multicolumn{2}{c|}{Neural Network Model} \\
      \cline{3-6}
      Method & Method for Data Collection & Position & Force & Input & Output \\
      \arrayrulecolor{black}
      \hline
      SM-w/o-Force            & Position symmetrical bilateral control & \checkmark & -- & Slave (6~dims.) & Master (6~dims.) \\
      SS-4CH           & 4ch bilateral control & \checkmark & \checkmark & Slave (9~dims.) & Slave (9~dims.) \\
      SM-4CH (Proposed method)            & 4ch bilateral control & \checkmark & \checkmark & Slave (9~dims.) & Master (9~dims.) \\
      \hline
    \end{tabular}
  \end{center}
\end{table*}
The experimental setup is shown in Fig.~\ref{fig:experiment}.
A spoon was affixed to the slave robot. The human manipulated a fork.
The task was serving several types of objects on a plate.
The objects on the plate were picked up through the cooperation between the human and robot and were served to the next plate.
Note that the position of the two plates was fixed on the desk.
In this study, comparative experiments of three types of imitation learning methods including the proposed method were verified.
The following three methods were validated in order to compare the effect of force control and variables to be inferred. A summary is shown in the Table~\ref{table:experiment}.
The three methods are as follows.
\begin{enumerate}
  \item SM-w/o-Force:\\
  The model was trained as a slave--master model based on position symmetrical bilateral control where force control in 4ch bilateral control is omitted.
  The slave--master model implies the LSTM model that infers the next states of the master robot from the current states of the slave robot.
  This method consisted of position control.
  The effects of force control were compared.
  \item SS-4CH:\\
  The model was trained as a slave--slave model based on 4ch bilateral control.
  The slave--slave model implies the LSTM model that infers the next states of the slave robot from the current states of the slave robot.
  This method consisted of position control and force control.
  This method is categorized in the same group of conventional imitation learning \cite{iit1}--\cite{force} that infer the next response values.
  The effects of inference of command values were compared.
  \item SM-4CH:\\
  The model was trained as a slave--master model based on 4ch bilateral control (proposed method). 
  This method consisted of position control and force control.
  Action force was measured and the dynamic interactions could be handled.
\end{enumerate}
\subsection{Data Collection}
Training data were generated using two types of bilateral control for comparative experiments.
Position symmetrical bilateral control was used for SM-w/o-Force. Conversely, 4ch bilateral control was used for SS-4CH and SM-4CH.
Conditions during the collection of the training data are shown on the left side of Fig.~\ref{fig:experiment}.
An operator manipulated the master robot while the slave robot served the food.
Snapshots of the slave robot's motion while collecting the training data are shown in Fig.~\ref{fig:salad_seq}.
In SM-w/o-Force, angle and angular velocity responses of the slave and master robots were recorded every 1~ms. Similarly, in SS-4CH and SM-4CH, angle, angular velocity, and torque responses of both the robots were recorded.
In the experiment, the four types of foods shown in the upper portion of Fig.~\ref{fig:food} were used to train serving tasks.
The trial time for a task was 7~s. A total of 20 trials were performed for each food. Therefore, 80 trials were performed for each type of bilateral control.

\subsection{Training Neural Network Model}
As shown in Table~\ref{table:experiment}, in SM-w/o-Force and SM-4CH, the response values of the slave robot were input to the neural network model, and the model output the response values of the master robot.
In SS-4CH, the response values of the slave robot were input to the neural network model, and the model output the response values of the slave robot.
In all the methods, the neural network models were learned to infer the next state from the current state of the robots.
Learning parameters were as follows: Optimization function: Adam \cite{adam}; Batch size: 100; Epoch: 1000. 
Learning took approximately 20 min with GPU calculation.
The computer used for this process had an Intel Core i7 CPU 32 GB memory, and an NVIDIA GTX 1080 Ti GPU.
\subsection{Task Execution}
Using the trained model, the cooperative work between the robot and human was verified.
Conditions during task execution are shown on the right side of Fig.~\ref{fig:experiment}.
In these experiments, adaptability to untrained objects was verified as well.
Moreover, robustness against environmental changes was verified.
The first validation was the change in the spoon length. 
Compared to the data collection phase, the spoon was fixed approximately 2~cm farther from the tip of the manipulator.
At this time, the position and height at which the robot was placed were adjusted according to the change in the length of the spoon.
This is because the relationship between the trajectory of the tip of the spoon and the position of the plate had to be kept the same as data collection phase.
This change is written as ``Change in the spoon length'' in Table~\ref{table:experimental_result}.
The second change is the height of fixing the plate.
The height of fixing the plate was raised approximately 2~cm above the data collection phase.
Note that the relationship between the trajectory of the spoon and the position of the plate was different from training.
This change is written as ``Change in the plate height'' in Table~\ref{table:experimental_result}.
\subsection{Experimental Results}\label{sec:result}
\begin{table*}[t]
  \begin{center}
    \caption{Success rates of cooperative work}
    \label{table:experimental_result}
    \begin{tabular}{|c|c||c|c|c|}
      \hline
      \multicolumn{2}{|c|}{} & \multicolumn{3}{c|}{Success Rate [\%]}\\
      \cline{3-5}
      \multicolumn{2}{|c|}{Method}& SM-w/o-Force & SS-4CH & SM-4CH (Proposed method) \\
      \hline
      & Salad 1& {\bf 100 (5/5)}& 60 (3/5)& {\bf 100 (5/5)}\\
      & Salad 2& {\bf 100 (5/5)}& 60 (3/5)& {\bf 100 (5/5)}\\
      & Pasta& 60 (3/5)& {\bf 100 (5/5)}& {\bf 100 (5/5)}\\
      Trained Objects& Orange& 60 (3/5)& 60 (3/5)& {\bf 80 (4/5)}\\
      \hline
      & Ping pong ball& 60 (3/5)& 60 (3/5)& {\bf 100 (5/5)}\\
      & Rubber ball& 40 (2/5)& 40 (2/5)& {\bf 100 (5/5)}\\
      & Towel handkerchief& 80 (4/5)& 40 (2/5)& {\bf 100 (5/5)}\\
      Untrained Objects& Sponge cake& 80 (4/5)& {\bf 100 (5/5)}& {\bf 100 (5/5)}\\
      \hline
      & Ping pong ball& 80 (4/5)& 60 (3/5)& {\bf 100 (5/5)}\\
      Chanege in the spoon length and Untrained Objects& Rubber ball& 60 (3/5)& 40 (2/5)& {\bf 100 (5/5)}\\
      \hline
      & Ping pong ball& 0 (0/5)& 0 (0/5)& {\bf 100 (5/5)}\\
      Change in the plate height and Untrained Objects& Rubber ball& 0 (0/5)& 0 (0/5)& {\bf 100 (5/5)}\\
      \hline
      \multicolumn{2}{|c||}{Total Success Rate [\%] }&   60.0 (36/60)& 51.7 (31/60)& {\bf 98.3 (59/60)}\\
      \hline
    \end{tabular}
  \end{center}
\end{table*}
\begin{figure}[t]
  \centering
  \includegraphics[width=8.5cm]{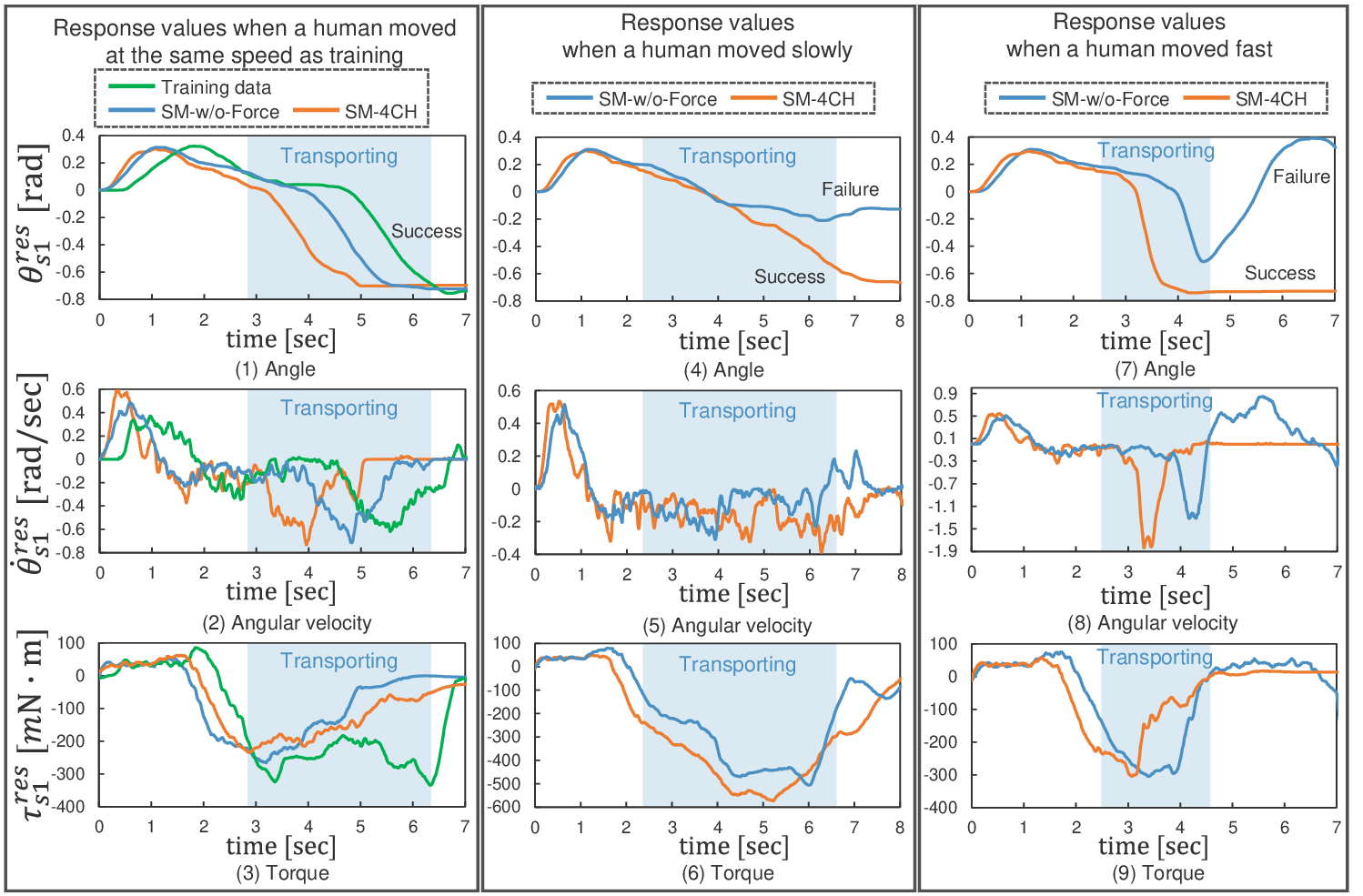}
  \caption{Result of SM-w/o-Force and SM-4CH}
  \label{fig:result position symmetry}
\end{figure}
In the proposed method, a series of movements from the task start to task completion was generated by one LSTM model.
LSTM switched each phase based on not only the time but changes in the response values of position, velocity, and force.
For example, the robot started the scooping action triggered by contact with the object and human at the scooping position.\par
The experimental results of the human--robot cooperative work for each method are shown in Table~\ref{table:experimental_result}.
Verification experiments were conducted five times for each object.
We defined success as the case where a robot scoops an object and places it on the next plate with human cooperation.
All objects had indefinite, completely different shape and rigidities.
Therefore, the robot was required to move robustly against the changes in shape, rigidity, and quantity of objects.\par
According to the experimental results, the proposed method (SM-4CH) showed the highest success rate.
Therefore, the proposed method is an effective approach in human--robot cooperation.
Most of the failure in the comparative methods occurred during transportation.
In SM-w/o-Force, failure occurred when the robot moved in the opposite direction to the next plate after scooping or stopped in the middle of transportation.
The response values of training data and the results of failures in autonomous operations are shown in Fig.~\ref{fig:result position symmetry}.
We focused on $\theta^{res}_{s1}$, $\dot{\theta}^{res}_{s1}$, and $\tau^{res}_{s1}$ of a joint that moved mainly during transportation.
The transportation is indicated by the blue area. Failure occurred in this area.
If a human moved at the same speed as the training data, the robot works well with SM-w/o-Force (on the left side in Fig.~\ref{fig:result position symmetry}). 
However, if a human moved at a different speed from the training data, the robot failed in the task (on the center and right side in Fig.~\ref{fig:result position symmetry}).
As the trajectory was generated by considering only the position information, the robot could not flexibly respond to human movement.
Fig.~\ref{fig:result position symmetry}. shows that the robot could respond to changes in human movement with SM-4CH.
Human motion is very complex and different in each trial.
Therefore, in SM-4CH, the robot could properly grasp an object using force control, and the appropriate motion was generated by adapting to the fluctuations in human motion.
The success rate of SS-4CH was the lowest in the three methods.
The master dataset included human skills to handle the dynamic interactions.
Therefore, the slave--master model can handle the dynamic interactions.
Nevertheless, it is difficult to consider the interactions in the case of the slave--slave model.
The difference between these models had a greater effect on the experimental results than the effect of the force control.\par
\begin{figure}[t]
  \centering
  \includegraphics[width=8.5cm]{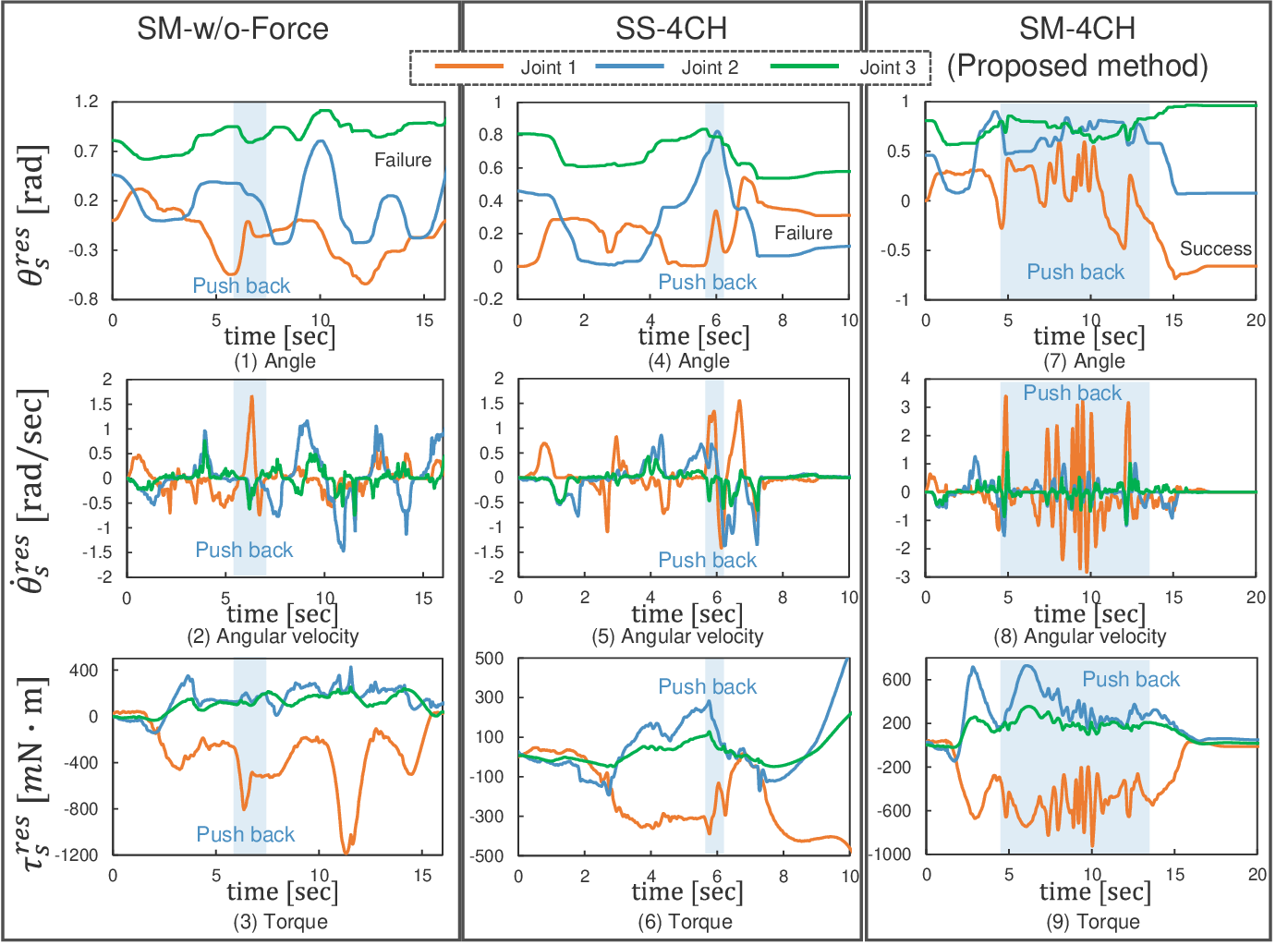}
  \caption{Results when pushed back by the human. The negative direction of $\theta^{res}_s$ is the transport direction, and the positive direction of $\theta^{res}_s$ is the pushed back direction. The robot was pushed back by the human many times, but eventually only the proposed method succeeded without dropping the object while maintaining proper gripping force.}
  \label{fig:pushback}
\end{figure}
Furthermore, in the proposed method, even if the human pushed back intentionally during transportation, the robot completed the tasks accurately. 
The response is shown in Fig.~\ref{fig:pushback}. No other comparison method was successful when pushed back.
In SM-w/o-Force, $\theta^{res}_{s2}$ and $\theta^{res}_{s3}$ did not work properly after being pushed back by the human, as shown in the comparison of Fig.~\ref{fig:pushback}-(1) and (7). The desired behavior was that $\theta^{res}_{s2}$ and $\theta^{res}_{s3}$ hardly change during transportation as shown in Fig.~\ref{fig:pushback}-(7) ({\it i.e.}, the height did not change during transportation). After the human pushed back once, the response was disturbed and the robot became impossible to continue the task with SM-w/o-Force.
In SS-4CH, after the human pushed back once, the robot moved in the opposite direction to the transport direction, and finally crashed into the plate and became unable to move.
In Fig.~\ref{fig:pushback}-(4), $\theta^{res}_{s1}$ moved in the opposite direction to the transport direction after the robot was pushed back.
This behavior was not the response of being pushed back by humans, but the robot moved by itself.
On the other hand, the proposed method completed the task, even if the human pushed back violently many times. 
The robot did not drop the object, and the response was not disturbed during transportation. 
This result indicated that the proposed method could respond appropriately to the irregular motion of the human owing to the force control and the proposed framework.
Details can be seen in the attached video and the video posted on YouTube (\url{https://youtu.be/duoairzAbh4}). At the end of the video, it was proved that the proposed method adapts to the fast motion of the human including the irregular motion.\par
In summary, the robot was able to execute tasks properly in response to changes in objects and environments with the proposed method.
Therefore, our framework based on 4ch bilateral control is an excellent approach towards human--robot cooperative work.
Note that the proposed method does not require any models of humans, special controllers for cooperation, {\it etc}.
Moreover, human--robot cooperation was realized only by imitation learning; special treatments are not required.\par

\section{CONCLUSION}
\label{sec:conclusion}
In this study, we verified task execution via cooperation between a human and robot.
The effectiveness of the proposed method in human--robot cooperation was verified through comparative experiments of three types of the methods without using force sensors.
The results of our method indicated the highest success rate compared to the other methods.
By using 4ch bilateral control, imitation learning of force control was enabled, and the robot could execute tasks requiring force adjustment.
Furthermore, the framework using the slave--master model was suggested to be suitable for imitation learning as per the comparative experiments between SS-4CH and SM-4CH.\par
Cooperative work succeeded in using the same framework as that of the conventional methods without special treatment.
That is to say, owing to our framework, the robot could adapt to complex human motion.
Therefore, the proposed method has great potential for realizing general manipulation. We believe that every task that can be realized by 4ch bilateral control can be realized by the proposed method.
In the future, our method may be applied to other tasks and it will be expanded using visual information.
\section*{ACKNOWLEDGMENT}
This work was supported by JST PRESTO Grant Number JPMJPR1755, Japan.

\end{document}